\documentclass{article}

\usepackage{microtype}
\usepackage{graphicx}
\usepackage{subfigure}
\usepackage{booktabs} 
\usepackage{amsmath}
\usepackage{bm}
\usepackage{natbib}
\usepackage{mathtools}
\usepackage{amssymb}
\usepackage{color,soul}
\usepackage{hyperref}

\DeclarePairedDelimiterX{\infdivx}[2]{(}{)}{%
  #1\;\delimsize\|\;#2%
}
\newcommand{\infdiv}{KL\infdivx}

\usepackage{hyperref}

\usepackage[accepted]{icml2019}

\icmltitlerunning{Variational Inference of Joint Models using Multivariate Gaussian Convolution Processes}

\begin{document}

\twocolumn[
\icmltitle{Variational Inference of Joint Models using Multivariate Gaussian Convolution Processes}

\icmlsetsymbol{equal}{*}

\begin{icmlauthorlist}
\icmlauthor{Xubo Yue}{umich}
\icmlauthor{Raed Kontar}{umich}
\end{icmlauthorlist}

\icmlaffiliation{umich}{Industrial \& Operations Engineering, University of Michigan, Ann Arbor, USA}
\icmlcorrespondingauthor{Raed Kontar}{alkontar@umich.edu}

\icmlkeywords{Machine Learning, ICML}

\vskip 0.3in
]

\printAffiliationsAndNotice{}  

\begin{abstract}
We present a non-parametric prognostic framework for individualized event prediction based on joint modeling of both longitudinal and time-to-event data. Our approach exploits a multivariate Gaussian convolution process (MGCP) to model the evolution of longitudinal signals and a Cox model to map time-to-event data with longitudinal data modeled through the MGCP. Taking advantage of the unique structure imposed by convolved processes, we provide a variational inference framework to simultaneously estimate parameters in the joint MGCP-Cox model. This significantly reduces computational complexity and safeguards against model overfitting. Experiments on synthetic and real world data show that the proposed framework outperforms state-of-the art approaches built on two-stage inference and strong parametric assumptions.
\end{abstract}

\section{Introduction}
\label{introduction}
In recent years, the multivariate Gaussian process (MGP) has drawn significant attention as an efficient non-parametric approach to predict longitudinal signal trajectories \cite{durichen2015multitask, moreno2018heterogeneous, kontar2018nonparametricb}. The MGP draws its roots from multitask learning where transfer of knowledge is achieved through a shared representation between training and testing signals. One neat approach that achieves this knowledge transfer, employs convolution processes to construct the MGP. Specifically, each signal is expressed as a convolution of latent functions drawn from a Gaussian process (GP). Commonalities amongst training and testing signals are then captured by sharing these latent functions across the outputs  \cite{titsias2010bayesian, alvarez2010efficient,alvarez2011computationally}. Consequently, the multiple signals can be expressed as a single output from a common multivariate Gaussian convolution process (MGCP). Indeed, many recent studies have demonstrated the MGCP ability to account for non-trivial commonalities in the data and provide accurate predictive results \cite{zhao2016variational, guarnizo2018fast, cheng2018multi}. 

\begin{figure}[H]
    \vskip -0.1in
    \centering
    \centerline{\includegraphics[width=\columnwidth]{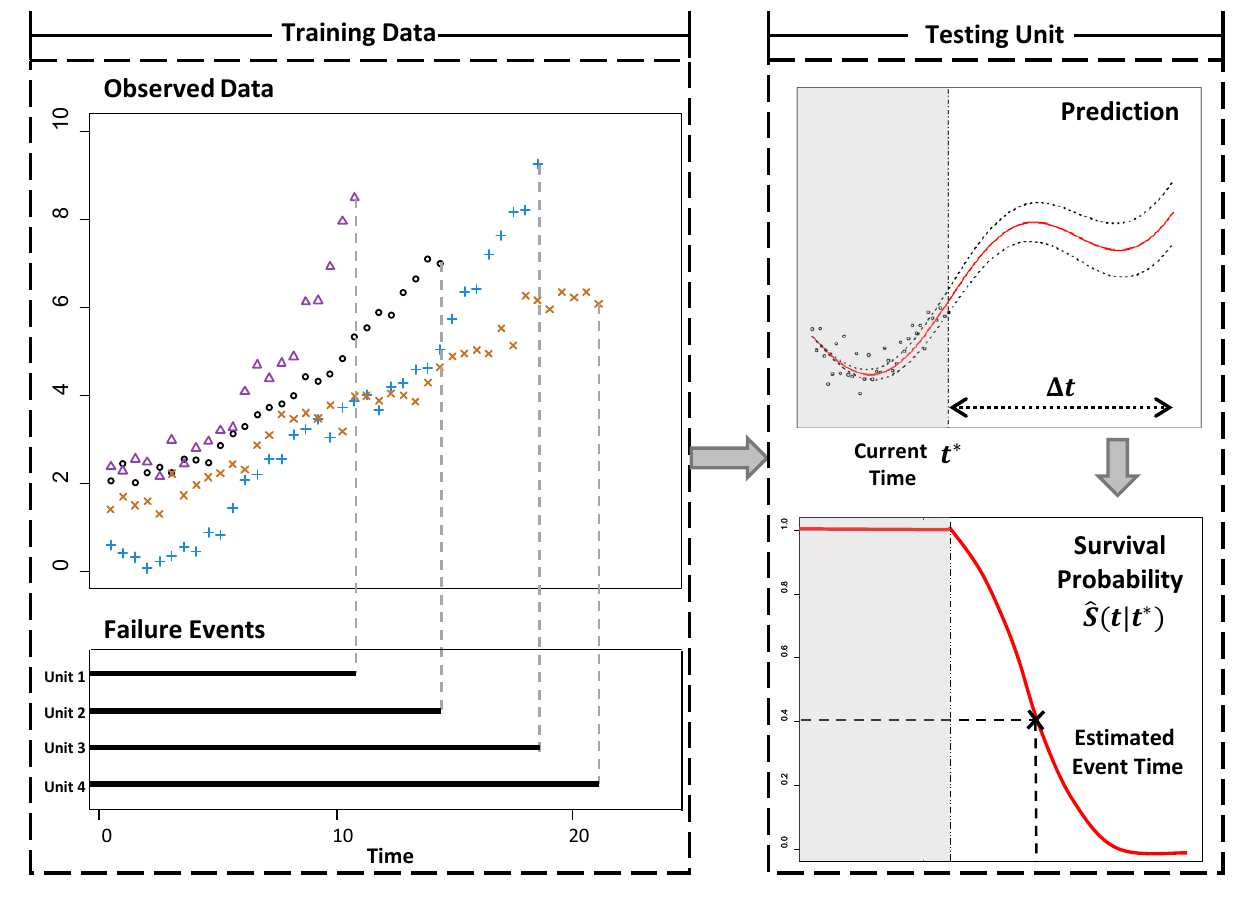}}
    \caption{Joint modeling of longitudinal and time-to-event data.}
    \label{framework}
    \vskip -0.2in
\end{figure}

In this article we exploit the MGCP to explore the following question: can we use both time-to-event data (also known as survival data) along with longitudinal signals to obtain a reliable event prediction? This is illustrated in Figure \ref{framework}. As shown in the figure, our goal is to utilize both survival data and longitudinal signals from training units to predict the survival probability of a partially observed testing unit. Naturally, the aforementioned question is often encountered in a wide range of applications, including: disease prognosis in clinical trials, event prediction using vital health signals from monitored patients at risk, remaining useful life estimation of operational units/machines and failure prognosis in connected manufacturing systems (e.g., nuclear power plants) \cite{tsiatis1995modeling, bycott1998comparison, gasmi2003general, pham2012machine, gao2015cloud, soleimani2018scalable}.

In order to link survival and longitudinal data, state-of-the-art methods have focused on joint models. The seminal work of Rizopoulos \cite{rizopoulos2011dynamic, rizopoulos2012joint} laid a foundation for joint models where a linear mixed effects model is used to model longitudinal signals. The coefficients of the mixed model are then used in a Cox model to compute the probability of event occurrence conditioned on the observed longitudinal signals. This idea provided the bases for many extensions and applications in the literature \cite{crowther2012flexible, wu2012analysis, zhu2012bayesian, crowther2013joint, proust2014joint, he2015simultaneous, rizopoulos2017dynamic, mauff2018joint}. It is important to note here that joint methods are in general built using a two-stage inference procedure. In two-stage inference, features from the longitudinal data are first learned, these estimated features are then inserted into a survival model to predict event probabilities. Indeed, many papers have shown that this two-stage procedure can produce competitive predictive results \cite{wulfsohn1997joint, yu2004joint, zhou2014remaining, mauff2018joint}. Nevertheless, the foregoing works are based on strong parametric assumptions where signals are assumed to follow a specific parametric form and all the signals (training and testing) exhibit that same functional form. In other words, signals behave according to a similar trend but at different rates (i.e., different parameter values). However, parametric methods are restrictive in many applications and if the specified form is far from the truth, predictive results will be misleading. Furthermore, the assumption that all signals possess the same functional form may not hold in real-life applications. For instance, units operated under different environmental conditions may exhibit different signal evolution rates and trends \cite{yan2016multiple, kontar2018nonparametrica}. Some recent efforts aimed to relax strong parametric assumptions using splines, continuous time Markov chains and the GP. Unfortunately, these methods still assume homogeneity across the population and focus on merely imputing the longitudinal data rather than predicting signal evolution within a  time interval of interest \cite{dempsey2017isurvive, soleimani2018scalable}. We here note that there has been some recent attempts at rebuilding the Cox model using a GP \cite{fernandez2016gaussian, Minyoung2018}. However these approaches are only based on survival data and do not handle joint modeling, which is the focus of this article.

To address the aforementioned challenges, we propose a flexible joint modeling approach denoted as MGCP-Cox. Our approach exploits the MGCP to model the evolution of longitudinal signals and a Cox model to map time-to-event data with longitudinal data modeled through MGCP. Event occurrence probability is then derived within any future interval $\Delta t$ as shown in Figure \ref{framework}. We also propose a variational inference framework using pseudo-inputs \cite{snelson2006sparse,damianou2013deep} to simultaneously estimate parameters in the joint MGCP-Cox model. This facilitates scalability to large data settings and safeguards against model overfitting. Finally, the advantageous features of the proposed method are demonstrated through numerical studies and a case study with real-world data in the application to finding the remaining useful lifetime of NASA Aero-propulsion engines.

The rest of the paper is organized as follows. In section 2 we briefly review survival analysis. In section 3, we present our joint modeling framework and the variational inference algorithm. Numerical experiments using synthetic data and real-world data are provided in section 4. Finally, section 5 concludes the paper with a brief discussion. A detailed code and the used real-world data are available in the \href{https://alkontar.engin.umich.edu/publications/}{supplementary materials}.

\section{Background: Survival Analysis}
\label{back:survival}
In this section, we will briefly review survival analysis which will be used for event prediction in the joint model. Survival analysis is a branch of statistics for analyzing time-to-event data and predicting the probability of occurrence of an event. For each individual unit $i$, the associated data is $\bm{\bm{\mathcal{D}}}_i=(V_i,\delta_i,\bm{Y}_i,\bm{w}_i)$, where $V_i=\min\{T_i,C_i\}$ is the event time (the unit failed at time $T_i$ or was censored at time $C_i$), $\delta_i\in\{0,1\}$ is an event indicator ($\delta_i=1/0$ indicates the unit has failed/censored), $\bm{Y}_i$ are the noisy observed longitudinal data (e.g., vital signals collected from patients) corresponding to the underlying latent values $\bm{f}_i$, and $\bm{w}_i$ is a set of time-invariant features (e.g., patient's gender). Typically, the continuous random variable $T_i$ is characterized by a survival function $S(t)=P(T\geq t)$ which represents the probability of survival up to time $t$. Another important term is the hazard function $h(t)=\lim_{\Delta\to 0}\frac{1}{\Delta}P(t<T\leq t+\Delta|T\geq t)=-\frac{d}{dt}\log S(t)$ and can be thought of as the instantaneous rate of occurrence of an event at time $t$. It is easy to show that $S(t)=\exp\{-\int_0^t h(u)du\}$. The term $\int_0^t h(u)du$ is called cumulative hazard function and is denoted by $H(t)$. The basic scheme of survival analysis is to find suitable models to explain relationships between the hazard function $h_i(t)$ and collected data $\bm{\mathcal{D}}_i$. These models are defined as survival models. 

Many survival models have been developed to analyze time-to-event data. They typically model the hazard function as a function of some time-varying and fixed features. One class of prevailed survival models is called the Cox model \cite{cox1972regression}, which has the form $h_i(t)=h_0(t)\exp[\bm{\gamma}^T\bm{w}_i+\beta f_i(t)]$, where $h_0(t)$ is a baseline hazard function shared by all individuals, and is typically modeled by the Weibull or a piecewise constant function, $\bm{\gamma}$ is a vector of coefficients for the fixed covariates (features), $f_i(t)$ is the feature estimated by a longitudinal model (e.g., linear mixed model, Gaussian Process), and $\beta$ is a scaling parameter for the time-varying covariates. Parameters in the Cox model are typically estimated by maximizing the log-likelihood function
\begin{equation}
\label{2:1}
    \begin{split}
        &\sum_{i=1}^N\log p(V_i,\delta_i|\bm{w}_i,\bm{f}_i)\\
        &=\sum_{i=1}^N\{\delta_i\log\big[h_0(V_i)\exp[\bm\gamma^T\bm{w}_i+\beta f_i(V_i)]\big]\\
        &-\int_0^{V_i}h_0(u)\exp[\bm\gamma^T\bm{w}_i+\beta f_i(u)]du\}.
    \end{split}
\end{equation}
For a comprehensive review of survival models, see \cite{kalbfleisch2011statistical}. Given an estimate of parameters from the Cox model, we can then obtain the event (failure) probability within a future time interval $\Delta t$ given the fact that the testing unit $i$ survives non-shorter than the current time instance $t^*$. This probability, denoted $\hat{P}_{\Delta t}$, is estimated as follows:
\vskip -0.2in
\begin{equation}
\label{2:3}
    \begin{split}
        &\hat{P}_{\Delta t}=1-\hat S(t^*+\Delta t|t^*,\bm{w}_i,\bm{f}_i)=1-\frac{\hat S(t^*+\Delta t|\bm{w}_i,\bm{f}_i)}{\hat S(t^*|\bm{w}_i,\bm{f}_i)}\\
        &=1-\exp\big\{-\int_{t^*}^{t^*+\Delta t}\hat h_0(u)\exp\big[\bm{\hat\gamma}^T\bm{w}_i+\hat\beta f_i(u)\big]du\big\},
    \end{split}
\end{equation}
where $\bm{w}_i$ and $\bm{f}_i$ are features for a testing unit $i$. Note that in Figure \ref{framework} we show the survival curve which is defined as $\hat{S}(t|t^\ast)=1-\hat{P}_{\Delta t}$ , where $t=t^*+\Delta t$.


\section{Joint Modeling and Variational Inference}
\label{model:joint}

\subsection{The multivariate Gaussian convolution process (MGCP)}
\label{3:mgp}

Assume data have been collected from $N$ units and let $\mathcal{I}=\{1,2,\ldots,N\}$ denote the set of all units. For unit $i$, its associated data is $\bm{\bm{\mathcal{D}}}_i=\{V_i,\delta_i, \bm{Y}_{i},\bm{w}_i\}$. The observed longitudinal signal is denoted by $\bm{Y}_{i}=(y_i(t_{i1}), \ldots, y_i(t_{il_i}))^T$, where $l_i$ represents the number of observations and $\{t_{ir}:r=1,\ldots,l_i\}$ denotes the inputs. We decompose the longitudinal signal as $y_i(t)=f_i(t)+\epsilon_i(t)$, where $f_i(\cdot)$ is a mean zero GP and $\epsilon_i(t)$ denotes additive noise with zero mean and $\sigma_\epsilon^2$ variance.

To obtain an accurate predictive result, we need to capture the intrinsic relatedness among $N$ signals. Particularly, we resort to the convolution process to model the latent function $f_i(t)$. We consider $K$ independent latent functions $\{X_k(t)\}_{k=1}^K$ and $NK$ different smoothing kernels $\{G_{i,k}(t):i\in \mathcal{I}\}_{k=1}^{K}$. The latent functions are assumed independent GPs with covariance $\mbox{cov}[X_k(t),X_{k}(t^{\prime})]=\kappa_k(t,t')$. We set  $G_{i,k}(t)=\alpha_{i,k}\mathcal{N}(0,\xi_{i,k}^{2})$ to be scaled Gaussian
kernels and $\kappa_k(t,t')$ to be squared exponential covariance functions \cite{alvarez2009sparse}.

\begin{figure}[!ht]
	\vskip 0.0in
	\begin{center}
		\centerline{\includegraphics[width=\columnwidth]{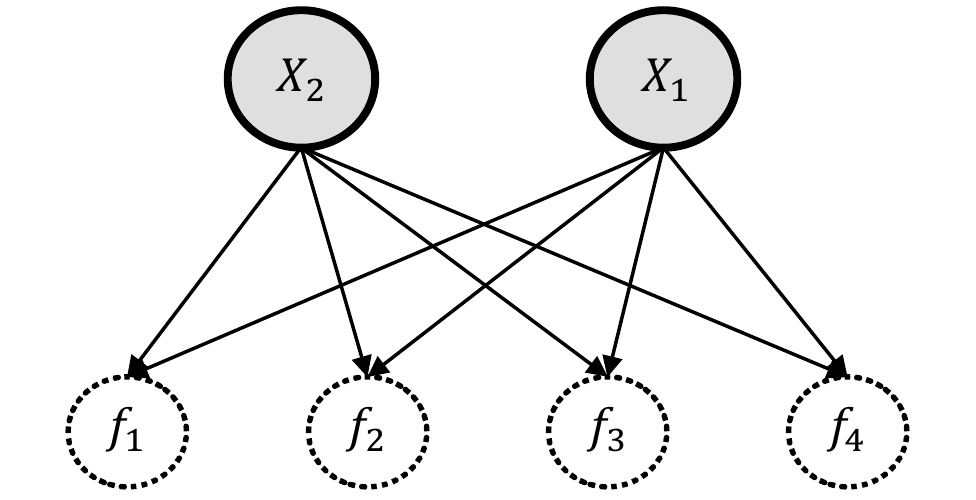}}
		\caption{Illustration of a convolution Process where two latent functions $X_1, X_2$ are shared across four outputs.}
		\label{convolution.pdf}
	\end{center}
	\vskip -0.5in
\end{figure}

\begin{equation}
\label{3:1}
    \begin{split}
        &\kappa_k(t,t')=\exp\bigg[-\frac{1}{2}\frac{(t-t')^2}{\lambda_k^{2}}\bigg]\\
        &=\sqrt{2\pi\lambda_k^{2}}\mathcal{N}(0,\lambda_k^{2})\coloneqq C_k\mathcal{N}(0,\lambda_k^{2}),
    \end{split}
\end{equation}

The GP $f_i(t)$ is then constructed by convolving the shared latent functions with the smoothing kernel as shown in (\ref{3:2}). This is the underlying principle  of the MGCP, where the latent functions $\{X_k(t)\}_{k=1}^K$ are shared across different outputs through the corresponding kernels $G_{i,k}(t)$. Since convolutions are linear operators on a function and since the latent function, a GP, is shared across multiple outputs then all outputs can be expressed as a jointly distributed GP, an MGCP. As shown in Figure \ref{convolution.pdf}, a key feature is that information is shared through different parameters encoded in the kernels $G_{i,k}(t)$. Outputs then can possess both shared and unique features. Thus, accounting for heterogeneity in the longitudinal data. 

\begin{equation}
\label{3:2}
    f_i(t)=\sum_{k=1}^K\int_{\mathbb{R}} G_{i,k}(t-u)X_k(u)du.
\end{equation}

Based on equation \eqref{3:2}, the covariance function between $f_i$ and $f_j$, and the covariance function between $f_i$ and $X_k$, can be calculated as
\begin{equation}
\label{3:3}
    \begin{split}
        &\kappa_{f_i,f_j}(t,t')\\
        &=\sum_{k=1}^K\int_{\mathbb{R}} G_{i,k}(t-u)\int_{\mathbb{R}} G_{j,k}(t'-u') \kappa_k(u,u')du'du\\
        &=\sum_{k=1}^K \alpha_{i,k}\alpha_{j,k}\sqrt{\frac{\lambda_k^{2}}{\eta^2_{i,j,k}}}\exp\Big(-\frac{1}{2}\frac{(t-t')^2}{\eta^2_{i,j,k}}\Big),\\
        &\kappa_{f_i,X_k}(t,u)=\int_{\mathbb{R}} G_{i,k}(t-u')\kappa_k(u,u')du'\\
        &=\alpha_{i,k}\sqrt{\frac{\lambda_k^{2}}{\eta^2_{i,k}}}\exp \Big(-\frac{1}{2}\frac{(t-u)^2}{\eta^2_{i,k}}\Big),
    \end{split}
\end{equation}
where $\eta^2_{i,j,k}=\xi_{i,k}^{2}+\xi_{j,k}^{2}+\lambda_k^{2}$ and $\eta^2_{i,k}=\xi_{i,k}^{2}+\lambda_k^{2}$.

Now denote the underlying latent values as $\bm{f}=\{\bm{f}_1^T,\ldots,\bm{f}_N^T\}^T$, where $\bm{f}_i=\{f_i(t_{i1}),\ldots,f_i(t_{il_i})\}^T$. The density function of $\bm{f}$ can be obtained as $p(\bm{f})=\mathcal{N}(\bm{0},\bm{K}_{\bm{f},\bm{f}})$, where $\bm{K}_{\bm{f},\bm{f}}$ sized $(\sum_{i=1}^N{l_i})\times(\sum_{i=1}^N{l_i})$ is the covariance function. The likelihood of $\bm{f}$ involves inverting the large matrix $\bm{K}_{\bm{f},\bm{f}}$. This operation has computational complexity $\mathcal{O}((\sum_{i=1}^N{l_i})^3)$ and storage requirement $\mathcal{O}((\sum_{i=1}^N{l_i})^2)$. To alleviate computational burden, we introduce $M$ pseudo-inputs from the latent functions denoted as  $\bm{X}_k(\bm{\mathcal{Z}})=[X_k(z_1),\ldots,X_k(z_M)]^T$ where $\bm{\mathcal{Z}}=\{z_i\}_{i=1}^M$. 

Since the latent functions are GPs, then any sample $\bm{X}_k(\bm{\mathcal{Z}})$ follows a multivariate Gaussian distribution. Conditioned on $\bm{X}_k(\bm{\mathcal{Z}})$, we next sample from the conditional prior $p(X_k(u)|\bm{X}_k(\bm{\mathcal{Z}}))$. In equation \eqref{3:2}, $X_k(u)$ can be approximated well by the expectation $\mathbb{E}(X_k(u)|\bm{X}_k(\bm{\mathcal{Z}}))$ as long as the latent functions are smooth \cite{alvarez2011computationally}. Denote by $\bm{X}=\{\bm{X}_1^T(\bm{\mathcal{Z}}),\ldots,\bm{X}_K^T(\bm{\mathcal{Z}})\}^T$. The probability distribution of $\bm{X}$ can be expressed as $p(\bm{X}|\bm{\mathcal{Z}})=\mathcal{N}(\bm{0},\bm{K}_{\bm{X},\bm{X}})$, where $\bm{K}_{\bm{X},\bm{X}}$ is a block-diagonal matrix such that each block is associated with the covariance of $X_k$ in (\ref{3:1}). By multivariate Gaussian identities, the probability distribution of $\bm{f}$ conditional on $\bm{X},\bm{\mathcal{Z}}$ is
\begin{equation}
\label{3:4}
    p(\bm{f}|\bm{X},\bm{\mathcal{Z}})=\mathcal{N}(\bm{K}_{\bm{f},\bm{X}}\bm{K}_{\bm{X},\bm{X}}^{-1}\bm{X},\bm{K}_{\bm{f},\bm{f}}-\bm{Q}),
\end{equation}
where $\bm{Q}=\bm{K}_{\bm{f},\bm{X}}\bm{K}_{\bm{X},\bm{X}}^{-1}\bm{K}_{\bm{X},\bm{f}}$. Therefore, $p(\bm{f})$ can be approximated by $p(\bm{f}|\bm{\mathcal{Z}})$, which is given as
\begin{equation}
\label{3:5}
    p(\bm{f}|\bm{\mathcal{Z}})=\int p(\bm{f}|\bm{X},\bm{\mathcal{Z}})p(\bm{X}|\bm{\mathcal{Z}}) d\bm{X}.
\end{equation}
By equation \eqref{3:5}, $p(\bm{Y})$ can then be approximated by $p(\bm{Y}|\bm{\mathcal{Z}})=\int p(\bm{Y}|\bm{f})p(\bm{f}|\bm{X},\bm{\mathcal{Z}})p(\bm{X}|\bm{\mathcal{Z}}) d\bm{f}d\bm{X}$.

\subsection{Joint Model and Variational Inference}
\label{3:vi}

Now following our convolution construction in \eqref{3:2}, the hazard function at time $t$ is given as
\begin{equation}
\label{3:13}
    \begin{split}
        h_i(t)=h_0(t)\exp\big[\bm{\gamma}^T\bm{w}_i+\beta \sum_{k=1}^K\int_{\mathbb{R}} G_{i,k}(t-u)X_k(u)du\big].
    \end{split}
\end{equation}
This key equation links the MGCP to the Cox model. We begin with presenting the log-likelihood of the joint model given observed data $\bm{\bm{\mathcal{D}}}=\{\bm{\bm{\mathcal{D}}}_i\}_{i=1}^N$. The marginal log-likelihood function is
\begin{equation}
\label{3:6}
    \begin{split}
        \log p(\bm{\mathcal{D}})&=\log \int p(\bm{\mathcal{D}}|\bm{f})p(\bm{f})d\bm{f}\\
        &\approx\log\int p(\bm{\mathcal{D}}|\bm{f})p(\bm{f}|\bm{\mathcal{Z}})d\bm{f}\\
        &=\int p(\bm{\mathcal{D}}|\bm{f})p(\bm{f}|\bm{X},\bm{\mathcal{Z}})p(\bm{X}|\bm{\mathcal{Z}})d\bm{X}d\bm{f}.
    \end{split}
\end{equation}
We would like to provide a good approximation of $\log p(\bm{\mathcal{D}})$ by introducing an evidence lower bound (ELBO) $\mathcal{L}$. This bound is calculated by finding the Kullback-Leibler (KL) divergence between the variational density $q(\bm{f},\bm{X}|\bm{\mathcal{Z}})$ and the true posterior density $p(\bm{f},\bm{X}|\mathcal{\bm{\mathcal{D}}},\bm{\mathcal{Z}})$. Specifically,
\begin{equation}
\label{3:7}
    \begin{split}
        &\infdiv{q(\bm{f},\bm{X}|\bm{\mathcal{Z}})}{p(\bm{f},\bm{X}|\bm{\mathcal{D}},\bm{\mathcal{Z}})}\\
        &=\int q(\bm{f},\bm{X}|\bm{\mathcal{Z}})\log\frac{q(\bm{f},\bm{X}|\bm{\mathcal{Z}})}{p(\bm{f},\bm{X}|\bm{\mathcal{D}},\bm{\mathcal{Z}})} d\bm{X}d\bm{f}\\
        &= \int q(\bm{f},\bm{X}|\bm{\mathcal{Z}})\log\frac{q(\bm{f},\bm{X}|\bm{\mathcal{Z}})p(\bm{\mathcal{D}})}{p(\bm{f},\bm{X},\bm{\mathcal{D}}|\bm{\mathcal{Z}})} d\bm{X}d\bm{f}\\
        &=\log p(\bm{\mathcal{D}})- \int q(\bm{f},\bm{X}|\bm{\mathcal{Z}})\log\frac{p(\bm{f},\bm{X},\bm{\mathcal{D}}|\bm{\mathcal{Z}})}{q(\bm{f},\bm{X}|\bm{\mathcal{Z}})}d\bm{X}d\bm{f}\\
        &=\log p(\bm{\mathcal{D}})-\mathcal{L}\geq 0.
    \end{split}
\end{equation}
The variaitonal density is assumed to be factorized as
\begin{equation}
\label{3:8}
    q(\bm{f},\bm{X}|\bm{\mathcal{Z}})=p(\bm{f}|\bm{X},\bm{\mathcal{Z}})q(\bm{X}).
\end{equation}
Maximizing the ELBO with respect to $q(\bm{X})$ and hyperparameters from the MGCP-Cox model can achieve purposes of variational inference and model selection simultaneously \cite{Minyoung2018}. By equation \eqref{3:7}, 
\begin{equation}
\label{3:9}
    \begin{split}
        \mathcal{L}&=\int q(\bm{f},\bm{X}|\bm{\mathcal{Z}})\log\frac{p(\bm{f},\bm{X},\bm{\mathcal{D}}|\bm{\mathcal{Z}})}{q(\bm{f},\bm{X}|\bm{\mathcal{Z}})}d\bm{X}d\bm{f}\\
        &=\int q_1(\bm{X})\int p(\bm{f}|\bm{X},\bm{\mathcal{Z}})\log p(\bm{\mathcal{D}}|\bm{f})d\bm{f}d\bm{X}\\
        &+\int q_2(\bm{X})\log\frac{p(\bm{X}|\bm{\mathcal{Z}})}{q(\bm{X})}d\bm{X}.
    \end{split}
\end{equation}
Furthermore, we can decompose $\log p(\bm{\mathcal{D}}|\bm{f})=\log p(\bm{Y}|\bm{f})+\log p(\bm{V},\bm{\delta}|\bm{w},\bm{f})$, where $\bm{V}=\{V_i\}_{i=1}^N$, $\bm{\delta}=\{\delta_i\}_{i=1}^N$ and $\bm{w}=\{w_i\}_{i=1}^N$. Based on equation \eqref{3:9}, the MGCP propagates uncertainties through the latent processes to the Cox model. 

It is desirable to find a closed form of the ELBO in equation \eqref{3:9}. Since $p(\bm{Y}|\bm{f})$ and $p(\bm{f}|\bm{X},\bm{\mathcal{Z}})$ are both Gaussian, we can obtain
\begin{equation}
\label{3:10}
    \begin{split}
        &\int p(\bm{f}|\bm{X},\bm{\mathcal{Z}})\log p(\bm{Y}|\bm{f})d\bm{f}\\
        &=\log\mathcal{N}(\bm{K}_{\bm{f},\bm{X}}\bm{K}_{\bm{X},\bm{X}}^{-1}\bm{X},\sigma_\epsilon^2\bm{I})-\frac{1}{2\sigma_\epsilon^2}\text{Tr}(\bm{K}_{\bm{f},\bm{f}}-\bm{Q}),
    \end{split}
\end{equation}
where $\text{Tr}(\cdot)$ is a trace operator. Therefore, the ELBO can be simplified as
\begin{align}
\label{3:11}
        \mathcal{L}&= -\frac{1}{2\sigma_\epsilon^2}\text{Tr}(\bm{K}_{\bm{f},\bm{f}}-\bm{Q}) \nonumber\\
        &+\int q_1(\bm{X})\log\frac{\mathcal{N}(\bm{K}_{\bm{f},\bm{X}}\bm{K}_{\bm{X},\bm{X}}^{-1}\bm{X},\sigma_\epsilon^2\bm{I})p(\bm{X}|\bm{\mathcal{Z}})}{q(\bm{X})}d\bm{X} \nonumber\\
        &+\int q_2(\bm{X})p(\bm{f}|\bm{X},\bm{\mathcal{Z}})\log p(\bm{V},\bm{\delta}|\bm{w},\bm{f})d\bm{f}d\bm{X}
\end{align}
We compute the optimal upper bound of $\mathcal{L}$ by reversing Jensen's inequality. This gives an optimal distribution $q_1^*(\bm{X})$ and
\begin{equation}
\label{3:12}
    \begin{split}
        &\mathcal{L}^*\\
        &=\log\int \mathcal{N}(\bm{K}_{\bm{f},\bm{X}}\bm{K}_{\bm{X},\bm{X}}^{-1}\bm{X},\sigma_\epsilon^2\bm{I})p(\bm{X}|\bm{\mathcal{Z}})d\bm{X}+\bm{PE}\\
        &\quad+\int q_{2}(\bm{X})p(\bm{f}|\bm{X},\bm{\mathcal{Z}})\log p(\bm{V},\bm{\delta}|\bm{w},\bm{f})d\bm{f}d\bm{X}\\
        &=\log[\mathcal{N}(\bm{0},\sigma_\epsilon^2 \bm{I}+\bm{Q})]+\bm{PE}\\
        &\quad+\int q_{2}(\bm{X})p(\bm{f}|\bm{X},\bm{\mathcal{Z}})\log p(\bm{V},\bm{\delta}|\bm{w},\bm{f})d\bm{f}d\bm{X},
    \end{split}
\end{equation}
where $\bm{PE}=-\frac{1}{2\sigma_\epsilon^2}\text{Tr}(\bm{K}_{\bm{f},\bm{f}}-\bm{Q})$. $\bm{PE}$ can be thought of as a penalization term that regularizes the estimation of the parameters. Note that the first two terms in equation \eqref{3:12} can be computed in $\mathcal{O}((\sum_{i=1}^N{l_i})M^2)$ \cite{snelson2006sparse}.

We will present a solution to solve the last integration in equation \eqref{3:12} in the following section.

\subsection{Variational Inference on Cox Model}
\label{3:cox}
Parameters in the Cox model can be attained by maximizing the following log-likelihood function:
\begin{equation}
\label{3:14}
    \begin{split}
        &\log p(\bm{V},\bm{\delta}|\bm{w},\bm{f})\\
        &=\sum_{i=1}^{N}\log p(V_i,\delta_i|\bm{w}_i,\bm{f}_i)=\sum_{i=1}^{N}\big\{\delta_i\log \big[h_0(V_i)\exp[\bm{\gamma}^T\bm{w}_i\\
        &+\beta \sum_{k=1}^K\int_{\mathbb{R}} G_{i,k}(V_i-u)X_k(u)du]\big]-\int_0^{V_i}h_0(u)\exp[\bm{\gamma}^T\bm{w}_i\\
        &\quad+\beta \sum_{k=1}^K\int_{\mathbb{R}} G_{i,k}(u-v)X_k(v)dv]du\big\}.
    \end{split}
\end{equation}

In equation \eqref{3:12}, we obtain the optimal $q_1^*(\bm{X})$ to maximize the ELBO. In this section, we will use  $q_1^*(\bm{X})$ to approximate $q_{2}({\bm{X}})$. 
Specifically, the optimal $q_1^*(\bm{X})$ has the form
\begin{equation}
\label{3:15}
    \begin{split}
        &q_1^*(\bm{X})\\
        &=\mathcal{N}(\sigma_\epsilon^{-2}\bm{K}_{\bm{X},\bm{X}}(\bm{K}_{\bm{X},\bm{X}}+\sigma_\epsilon^{-2}\bm{K}_{\bm{X},\bm{f}}\bm{K}_{\bm{f},\bm{X}})^{-1}\bm{K}_{\bm{X},\bm{f}}\bm{Y},\\
        &\bm{K}_{\bm{X},\bm{X}}(\bm{K}_{\bm{X},\bm{X}}+\sigma_\epsilon^{-2}\bm{K}_{\bm{X},\bm{f}}\bm{K}_{\bm{f},\bm{X}})^{-1}\bm{K}_{\bm{X},\bm{X}})\coloneqq\mathcal{N}(\bm{m},\bm{s}).
    \end{split}
\end{equation}
It is easy to show that $q(\bm{f}|\bm{\mathcal{Z}})$ has the normal distribution with parameter $\bm{\mu},\bm{\Sigma}$. Specifically,
\begin{equation}
\label{3:16}
    \int q_1^*(\bm{X})p(\bm{f}|\bm{X},\bm{\mathcal{Z}})d\bm{X}=q(\bm{f}|\bm{\mathcal{Z}})\coloneqq q(\bm{f})\sim\mathcal{N}(\bm{\mu},\bm{\Sigma}),
\end{equation}
where 
\begin{equation*}
\label{3:17}
    \begin{split}
        \bm{\mu}&=\bm{K}_{\bm{f},\bm{X}}\bm{K}_{\bm{X},\bm{X}}^{-1}\bm{m},\\
        \bm{\Sigma}&=\bm{K}_{\bm{f},\bm{f}}-\bm{K}_{\bm{f},\bm{X}}\bm{K}_{\bm{X},\bm{X}}^{-1}(\bm{I}-\bm{s}\bm{K}^{-1}_{\bm{X},\bm{X}})\bm{K}_{\bm{X},\bm{f}}.
    \end{split}
\end{equation*}
The last integration in equation \eqref{3:12} can be simplified to
\begin{equation}
\label{3:18}
    \begin{split}
        &\int q(\bm{f})\log p(\bm{V},\bm{\delta}|\bm{w},\bm{f})d\bm{f}\\
        &=\int q(\bm{f})\sum_{i=1}^N\log p(V_i,\delta_i|\bm{w}_i,\bm{f}_i)d\bm{f}\\
        &=\int q(\bm{f})\sum_{i=1}^N\big\{\delta_i\log \big[h_0(V_i)\exp[\bm{\gamma}^T\bm{w}_i+\beta f_i(V_i)]\big]\\
        &-\int_0^{V_i}h_0(u)\exp[\bm{\gamma}^T\bm{w}_i+\beta f_i(u)]du\big\}d\bm{f}.
    \end{split}
\end{equation}
The first term in equation \eqref{3:18} can be calculated analytically. For each unit $i$,
\begin{equation}
\label{3:19}
    \begin{split}
        &\int q(\bm{f})\delta_i\log [h_0(V_i)\exp[\bm{\gamma}^T\bm{w}_i+\beta f_i(V_i)]]d\bm{f}\\
        &=\delta_i\big\{\log h_0(V_i)+\bm{\gamma}^T\bm{w}_i+\beta\int q(\bm{f}) f_i(V_i)d\bm{f}\big\}\\
        &=\delta_i\big\{\log h_0(V_i)+\bm{\gamma}^T\bm{w}_i+\beta\mathbb{E}_{q(\bm{f})}[f_i(V_i)]\big\}\\
        &=\delta_i\big\{\log h_0(V_i)+\bm{\gamma}^T\bm{w}_i+\beta\mu_i(V_i)\big\},
    \end{split}
\end{equation}
where $\mathbb{E}_{q(\bm{f})}[f_i(V_i)]=\bm{K}_{f_i(V_i),\bm{X}}\bm{K_{X,X}}^{-1}\bm{m}\coloneqq\mu_i(V_i).$ In the last step, we applied the Fubini's theorem to interchange integrals. The second term in equation \eqref{3:18} can be estimated by the moment generating function (MGF) and the numerical integration. For each unit $i$, 
\begin{equation}
\label{3:20}
    \begin{split}
        &\int q(\bm{f})(-\int_0^{V_i}h_0(u)\exp[\bm{\gamma}^T\bm{w}_i+\beta f_i(u)]du)d\bm{f}\\
        &=-\int_0^{V_i}h_0(u)\exp[\bm{\gamma}^T\bm{w}_i]\int q(\bm{f})\exp[\beta f_i(u)]d\bm{f}du\\
        &=-\int_0^{V_i}h_0(u)\exp[\bm{\gamma}^T\bm{w}_i]\exp\big[\beta [\mu_i(u)+\frac{1}{2}\sigma_i^2(u)]\big]du,
    \end{split}
\end{equation}
where $\mu_i(u)\coloneqq\bm{K}_{f_i(u),\bm{X}}\bm{K_{X,X}}^{-1}\bm{m}$, and $\sigma_i^2(u)\coloneqq\bm{K}_{f_i(u),f_i(u)}-\bm{K}_{f_i(u),\bm{X}}\bm{K}_{\bm{X},\bm{X}}^{-1}(\bm{I}-\bm{s}\bm{K}^{-1}_{\bm{X},\bm{X}})\bm{K}_{\bm{X},f_i(u)}$.
We can assume $h_0(t)$ to be an exponential function $\exp(b+\psi(t-\min\{V_i\}_{i=1}^N))$, where $b$,$\psi$ are parameters to be learned and $h_0(t)=0$ when $t<\min\{V_i\}_{i=1}^N$ because units are not subject to risk before the first failure event. Note that the baseline hazard is non-decreasing with time and thus we add one constraint $\psi\in\mathbb{R}_+$. To obtain a good baseline hazard prediction given the estimated $\hat b,\hat \psi$, we can calculate the cumulative hazard at time point $t$ as $H(t)=\sum_{u\in\mathcal{F}(t)}\hat h_0(u),\forall t$, where $\mathcal{F}(t)\coloneqq\big\{\{V_{(1)},V_{(2)},\ldots,V_{(N-1)}\}\cup\{0,1,2,\ldots,V_{(N)}\}\big\}\cap[0,t]$, and $V_{(i)}$ is the $i$-th smallest element in $\{V_i\}_{i=1}^N$. Then we fit a regularized smooth spline to $H(t)$ \cite{ruppert2002selecting}. The predicted baseline hazard at $u\in[t^*,t^*+\Delta t^*]$ can be estimated by $\frac{d\hat H(t)}{dt}\Big|_{t=u}$ \cite{rosenberg1995hazard}.       

The $\mathcal{L}^*$ is maximized with respect to the parameters $\Theta=(\bm{\theta}$, $\sigma_\epsilon$, $\bm{\gamma}$, $\beta$, $b$, $\psi)$, where $\bm{\theta}=(\{\lambda_k,\xi_{i,k},\alpha_{i,k}\}_{i=1,k=1}^{N,K})$, by the gradient-based method. Specifically, We can obtain the optimal parameters $\hat\Theta$ by maximizing $\mathcal{L}^*$ subject to $\psi\geq 0$.

\subsection{Event Prediction}
\label{3:event}

Without loss of generality, we focus on predicting the event occurrence probability for unit $N$. Suppose observations from the testing unit $N$ have been collected up to time $t^*$. The survival model computes the event probabilities conditioned on the predicted longitudinal features $\bm{f}_N(u),u\in[t^*,t^*+\Delta t]$. Given estimated parameters, and following \eqref{2:3}, we are interested in calculating
\begin{equation}
\label{3:22}
    \begin{split}
        &1-\hat S(t^*+\Delta t|t^*,\bm{w}_N,\bm{f}_N)=1-\frac{\hat S(t^*+\Delta t|\bm{w}_N,\bm{f}_N)}{\hat S(t^*|\bm{w}_N,\bm{f}_N)}\\
        &=1-\exp\{-\int_{t^*}^{t^*+\Delta t}\hat h_0(u)\exp\big[\bm{\hat\gamma}^T\bm{w}_N+\hat\beta f_N(u)\big]du\}.
    \end{split}
\end{equation}
Based on equation \eqref{3:22}, the accurate extrapolation within $\Delta t$ is essential. In the MGCP, the predictive distribution for any new input point $T$ is given by
\begin{equation}
\label{3:23}
    \begin{split}
        &p(f_N(T^*)|\bm{Y})=\int p(f_N(T^*)|\bm{X})p(\bm{X}|\bm{Y})d\bm{X}\\
        &=\int \mathcal{N}(\bm{K}_{f_N(T^*),\bm{X}}\bm{K}_{\bm{X},\bm{X}}^{-1}\bm{X},\bm{W})p(\bm{X}|\bm{Y})d\bm{X}\\
        &=\int \mathcal{N}(\bm{K}_{f_N(T^*),\bm{X}}\bm{K}_{\bm{X},\bm{X}}^{-1}\bm{X},\bm{W})\frac{p(\bm{Y}|\bm{X})p(\bm{X})}{p(\bm{Y})}d\bm{X}\\
        &=\mathcal{N}(\bm{AD}^{-1}\bm{Y}, \bm{K}_{f_N(T^*),f_N(T^*)}-\bm{AD}^{-1}\bm{A}^T),
    \end{split}
\end{equation}
where $\bm{A}=\bm{K}_{f_N(T^*),\bm{X}}\bm{K}_{\bm{X},\bm{X}}^{-1}\bm{K}_{\bm{X},\bm{f}}$, $\bm{D}=\bm{Q}+\sigma_\epsilon^2\bm{I}$ and $\bm{W}=K_{f_N(T^*),f_N(T^*)}-\bm{K}_{f_N(T^*),\bm{X}}\bm{K}_{\bm{X},\bm{X}}^{-1}\bm{K}_{\bm{X},f_N(T^*)}$. We have used $K_{f_N(T^*),f_N(T^*)}$ as a notation to indicate when the covariance matrix is evaluated at the $T^*$. Consequently, the predicted signal at the time point $T^*$ for unit $N$ is $\hat{f}_N(T^*)=\bm{AD}^{-1}\bm{Y}$.


\section{Experiments}
\label{experiment}
\begin{figure*}[!htb]
    \vskip 0.0in
    \begin{center}
    \centerline{\includegraphics[width=1.3\columnwidth]{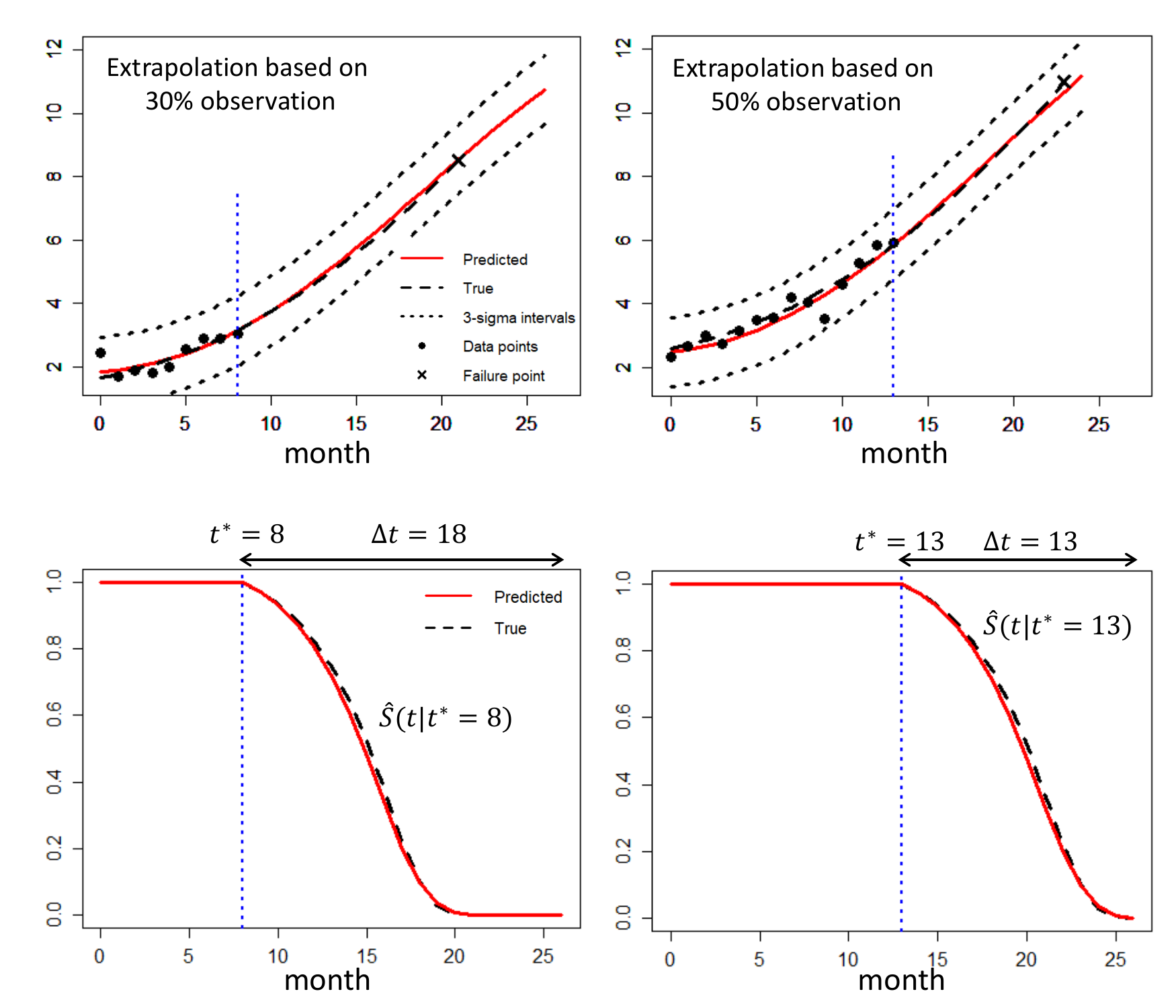}}
    \caption{Results of extrapolations and survival probabilities prediction for two testing units. Figures in the first row show results of extrapolations based on different observation percentiles. Figures in the second row show true and predicted survival curves.}
    \label{extrapolation}
    \end{center}
    \vskip -0.2in
\end{figure*}

We conduct case studies to demonstrate the performance of our proposed methodology, denoted as MGCP-Cox. Both synthetic and real-world data are used. We also provide an illustrative example in Figure \ref{extrapolation} to demonstrate the benefits of the MGCP-Cox model.

\subsection{Data setting}
\label{General setting for case studies}
For the synthetic data we assume that the underlying true path for unit $i$ has the form $y_i(t)=\bm{z}^T(t)\bm{b}_i+\epsilon_i(t)=b_{i0}+b_{i1}t+b_{i2}t^2+\epsilon_i(t)$, where $\epsilon_i(t)\sim\mathcal{N}(0, 0.1)$, $\bm{z}^T(t)=[1,t,t^2]$ and $\bm{b}_i=[b_{i0}, b_{i1}, b_{i2}]^T\sim\mathcal{N}(\bm{\mu}_b,\bm{\Sigma}_b)$ with $\bm{\mu}_b=[2.5, 0.1, a]^T$ and $\bm{\Sigma}_b=\begin{bmatrix}0.2&-4e-4&-8e-5\\-4e-4&3e-6&3e-7\\-e-5&3e-7&1e-7\end{bmatrix}$ where $a\sim \mbox{uniform}(0.003,0.03).$  Without loss of generality, we assume that the time unit is month and that signals were obtained regularly at each month up to their failure or censoring time.  An example of the signals is shown in the top row of Figure \ref{extrapolation}. For each unit we specify a time-invariant feature $w_i\in\{0,1\}$ generated by a Bernoulli distribution with $p=0.5$. In the Cox model, we use the Weibull baseline hazard rate function $h_0(t)=\lambda\rho t^{\rho-1}$ with $\lambda=0.001$ and $\rho=1.05$.  We generate the failure time $T_i$ for each unit by rejection sampling using its probability density function $h_i(t)S_i(t)$ . We set $\gamma=0$ and $\beta=0.5$. Also, we randomly select $5\%$ of the units to be right censored. The number of units generated is $N=20$ and the experiment is repeated for $Q=100$ times. Detailed code for data generation is provided in the \href{https://alkontar.engin.umich.edu/publications/}{supplementary materials}. 

For the real-world case study we use the C-MAPSS dataset provided by the National Aeronautics and Space Administration (NASA).  The dataset contains failure time data of aircraft turbofan engines and degradation signals from informative sensors mounted on these engines. Note that in our analysis we standardize all sensor data. We refer readers to \citet{saxena2008} and \citet{liu2013data} for more details about the data. The C-MAPSS data publically available at: \url{https://ti.arc.nasa.gov/tech/dash/groups/pcoe/prognostic-data-repository/}.



\subsection{Baselines and Evaluations}
\label{base}
\begin{figure*}[!htb]
\centering
\vskip 0.0in
\begin{center}
\centerline{\includegraphics[width=1.7\columnwidth]{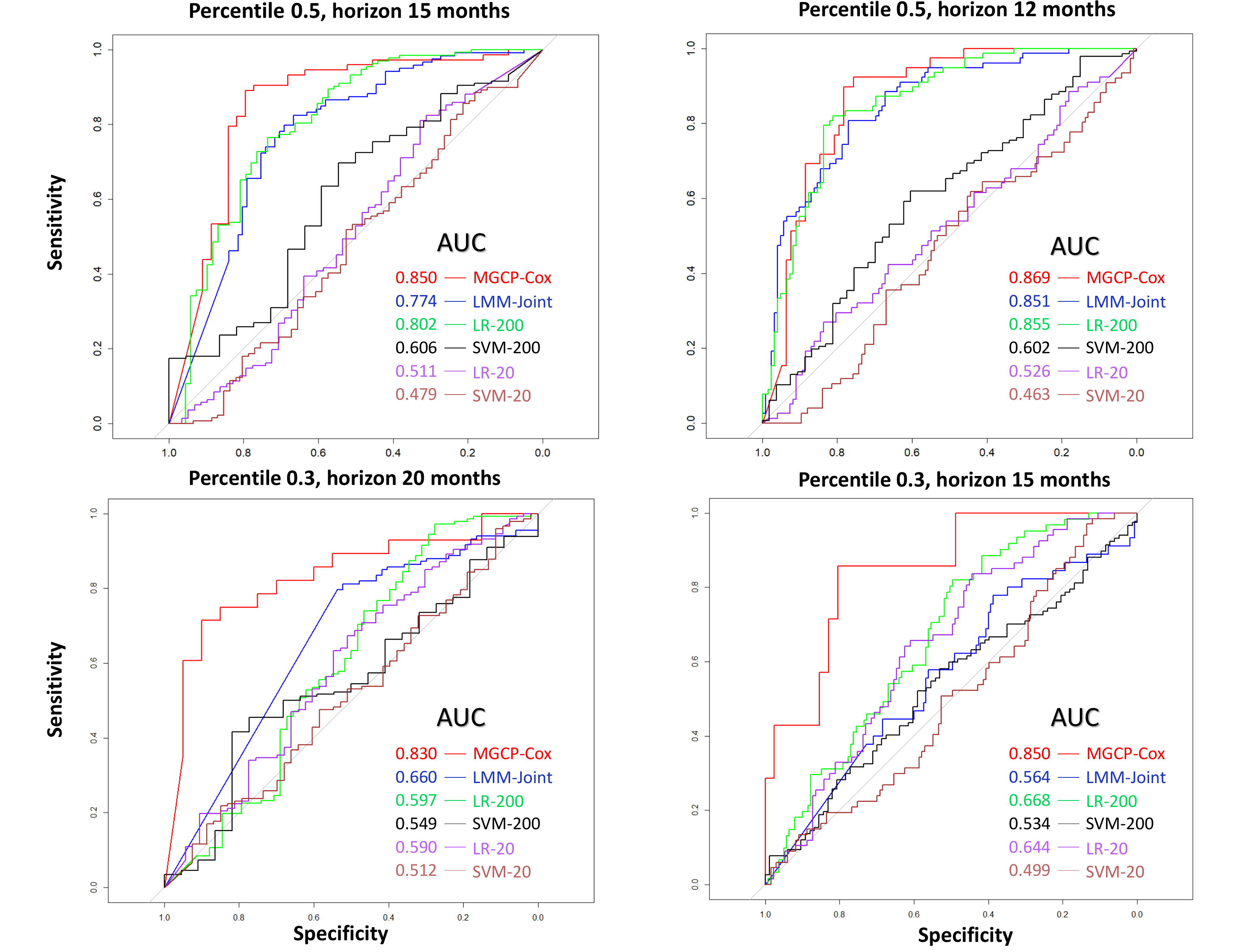}}
\caption{ROC curves from simulation studies under different percentile of observation $\alpha$.}
\label{roc}
\end{center}
\vskip -0.2in
\end{figure*}

We focus on predicting the event probability within a future time interval $\Delta t$. We consider $\Delta t=12,15,20$ months in this simulation study. Prediction performance at varying time points $t^*$ for the partially observed unit $N$ is then reported. The time instant $t^*=\alpha T_N$ is defined as the $\alpha$-observation percentile, where $T_N$ is the failure time of unit $N$. The values of $\alpha$ are specified as $30\%$, $50\%$. Figure \ref{extrapolation} shows some examples of units observed up to different percentiles of their failure time. Further, in our simulation studies, we benchmark our method with three other reference methods for comparison: (1) Logistic Regression
(LR) classifier: in the LR, event data is transformed into binary labels $\delta_i=1/0$ denoting whether units failed or not within the time interval $[t^{\ast},\Delta t+t^*]$. The time-fixed covariate $w_i$ and the last observed signal measurement at $t^{\ast}$ are used as the model predictors (2) Support Vector Machine (SVM) classifier: the SVM here is used as a flexible alternative to the LR. We use the radial basis kernel and determine parameters using $2$-fold cross-validation on the training data (3) Parametric Joint Model (LMM-Joint): we implement a state-of-the-art joint modeling algorithm using the linear mixed-effect model. The LMM-joint uses a general polynomial function whose corresponding degree is determined through an Akaike information criteria to model the signal path. Note that this framework estimates parameters from the mixed-effect model and the Cox model separately \cite{rizopoulos2011dynamic, zhou2014remaining, mauff2018joint}. Regarding our MGCP-Cox model we set the number of pseudo-inputs to $M=10$ and the number of latent functions to $K=1$. This setting is a commonly used setting for the MGCP \cite{alvarez2011computationally, zhao2016variational}. The performance of each method is then assessed by the Receiver Operating Characteristic (ROC) curve, which is a common diagnostic tool for binary classifier. The ROC curve is created by plotting the true positive rate (TPR) against the false positive rate (FPR). Predictive accuracy is then assessed through the area under the curve (AUC). The results from the synthetic data are shown in Figure \ref{roc}. Due to poor performance of both the LR and SVM on $N=20$, we also checked whether they can produce comparable results to the MGCP-Cox when $N=200$. We denote those models as LR-200 and SVM-200.

For the real data, the true survival probabilities are not available since we do not have information about the underlying parameters used to generate the data. Therefore, to evaluate model performance, we calculate the mean remaining lifetime of the testing unit, which is defined as $\widehat {mrl}(t^*)=\int_{t^*}^\infty \hat S(u|t^*,\bm{w}_N,\bm{f}_N)du$. This integration can be obtained by the Gauss-Legendre quadrature. The performance is assessed by the absolute error $AE=|{rl}_j-\widehat{mrl}_j|$ where ${rl}_j$ is the true remaining lifetime of the testing unit. We then report the distribution of the errors across all units using the boxplot in Figure \ref{box}. Similar to the synthetic data we use 30\% and 50\% percentiles to assess prediction accuracy. We also note  than we cannot obtain $\widehat {mrl}$ estimates from the SVM and LR as they transform event prediction into a time series classification problem. Therefore, only results from LMM-joint and MGCP-Cox are reported in Figure \ref{box}.

\subsection{Results}
\label{results}

The illustrative example in Figure \ref{extrapolation} demonstrates the behavior of our method. As shown in the figure, our joint model framework can provide accurate predictions of both longitudinal signals and event probabilities. The unique smoothing kernel $G_{i,k}$ for each individual allows flexibility in the prediction, since it enables each training signal to have its own characteristics. This substantiates the strength of the MGCP. Equipped with the shared latent processes, the model can infer the similarities among all units, and predict signal trajectory by borrowing strength from training units. 

\begin{figure}[!htb]
    \vskip 0.0in
    \begin{center}
    \centerline{\includegraphics[width=0.8\columnwidth]{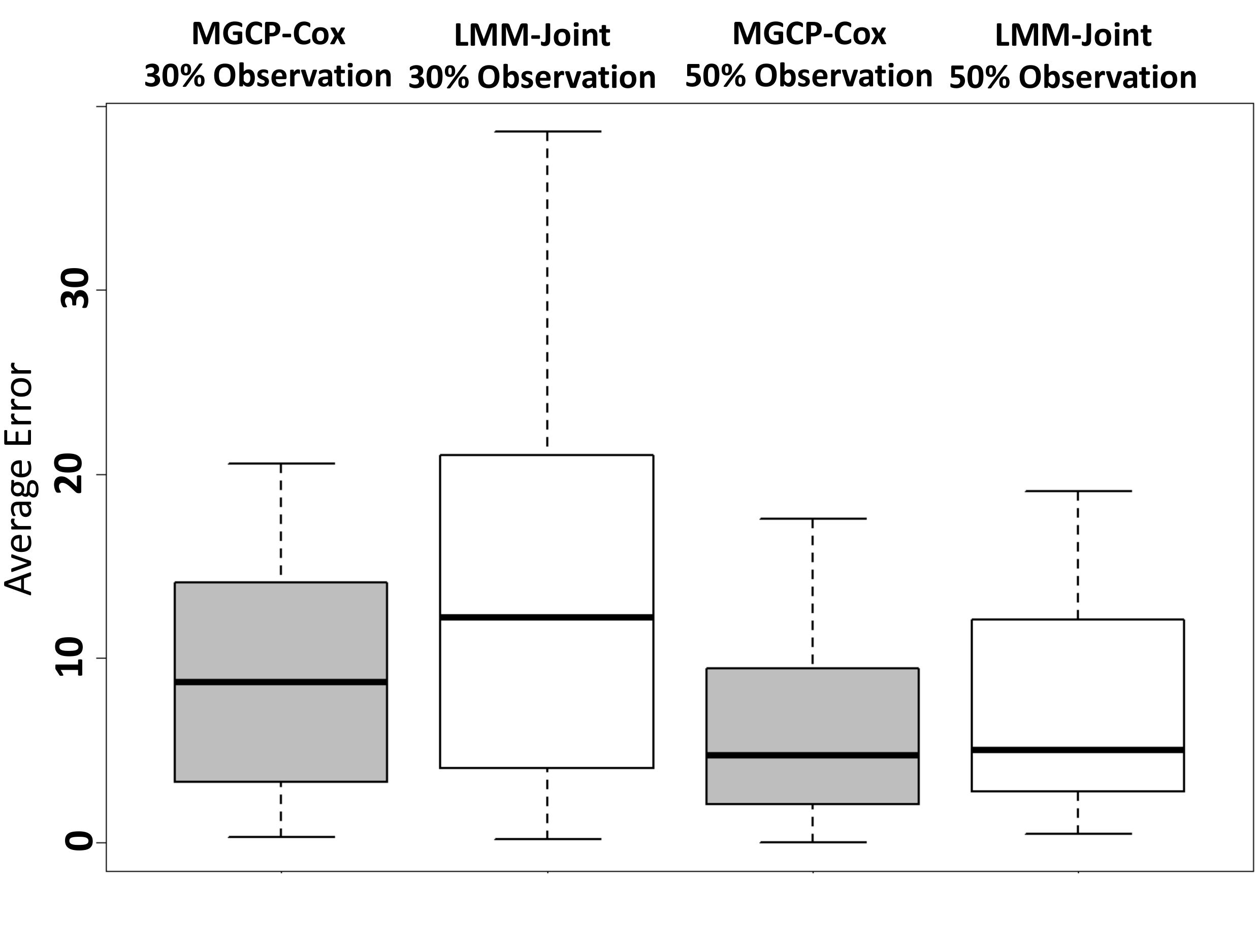}}
    \caption{Remaining life prediction accuracy from NASA data.}
    \label{box}
    \end{center}
    \vskip -0.4in
\end{figure}

The results in Figures \ref{roc} and \ref{box} indicate that our MGCP-Cox model clearly outperforms the benchmarked models. Based on the figure we can obtain some important insights. First, as expected, prediction errors significantly decrease as the lifetime percentiles increase. Thus, the prediction accuracy from the MGCP-Cox model will become more accurate as $t^{\ast}$ increases and more data are collected from an online monitored unit. Second, the prediction accuracy slightly decreases as we predict over a longer horizon (i.e. prediction is better for the near future). This is intuitively understandable as accuracy might decrease when predicting over a large region where not many training data might be observed. Third, the results show that the MGCP-Cox clearly outperforms LMM-joint. This result highlights the danger of parametric modeling and demonstrates the ability of our non-parametric approach to avoid model misspecifications. Fourth, even when the LR and SVM had a much larger number of units, the MGCP-Cox was still superior. This observation, also true to the LMM-Joint, highlights the strength of joint models. Lastly,  one striking feature, shown in Figures \ref{extrapolation}, \ref{roc} and \ref{box}, is that even with a small number of observations ($30\%$ observation percentile) from the testing unit we were still able to get accurate predictive results. This crucial in many applications as its allows early prediction of an event occurrence such as a disease or machine failure.

\section{Conclusion}
\label{discuss}

We have presented a flexible and efficient non-parametric joint modeling framework for longitudinal and time-to-event data. A variational inference framework using pseudo-inputs is established to jointly estimate parameters from the MGCP-Cox model. Empirical studies highlight the advantageous features of our model to predict signal trajectories and provide reliable event prediction. 

\bibliography{example_paper}

\begin{thebibliography}{42}
\providecommand{\natexlab}[1]{#1}
\providecommand{\url}[1]{\texttt{#1}}
\expandafter\ifx\csname urlstyle\endcsname\relax
  \providecommand{\doi}[1]{doi: #1}\else
  \providecommand{\doi}{doi: \begingroup \urlstyle{rm}\Url}\fi

\bibitem[{\'A}lvarez \& Lawrence(2009){\'A}lvarez and
  Lawrence]{alvarez2009sparse}
{\'A}lvarez, M. and Lawrence, N.~D.
\newblock Sparse convolved gaussian processes for multi-output regression.
\newblock In \emph{Advances in neural information processing systems}, pp.\
  57--64, 2009.

\bibitem[{\'A}lvarez \& Lawrence(2011){\'A}lvarez and
  Lawrence]{alvarez2011computationally}
{\'A}lvarez, M. and Lawrence, N.~D.
\newblock Computationally efficient convolved multiple output gaussian
  processes.
\newblock \emph{Journal of Machine Learning Research}, 12\penalty0
  (May):\penalty0 1459--1500, 2011.

\bibitem[{\'A}lvarez et~al.(2010){\'A}lvarez, Luengo, Titsias, and
  Lawrence]{alvarez2010efficient}
{\'A}lvarez, M., Luengo, D., Titsias, M., and Lawrence, N.~D.
\newblock Efficient multioutput gaussian processes through variational inducing
  kernels.
\newblock In \emph{Proceedings of the Thirteenth International Conference on
  Artificial Intelligence and Statistics}, pp.\  25--32, 2010.

\bibitem[Bycott \& Taylor(1998)Bycott and Taylor]{bycott1998comparison}
Bycott, P. and Taylor, J.
\newblock A comparison of smoothing techniques for cd4 data measured with error
  in a time-dependent cox proportional hazards model.
\newblock \emph{Statistics in medicine}, 17\penalty0 (18):\penalty0 2061--2077,
  1998.

\bibitem[Cheng(2018)]{cheng2018multi}
Cheng, C.
\newblock Multi-scale gaussian process experts for dynamic evolution prediction
  of complex systems.
\newblock \emph{Expert Systems with Applications}, 99:\penalty0 25--31, 2018.

\bibitem[Cox(1972)]{cox1972regression}
Cox, D.~R.
\newblock Regression models and life-tables.
\newblock \emph{Journal of the Royal Statistical Society: Series B
  (Methodological)}, 34\penalty0 (2):\penalty0 187--202, 1972.

\bibitem[Crowther et~al.(2012)Crowther, Abrams, and
  Lambert]{crowther2012flexible}
Crowther, M.~J., Abrams, K.~R., and Lambert, P.~C.
\newblock Flexible parametric joint modelling of longitudinal and survival
  data.
\newblock \emph{Statistics in Medicine}, 31\penalty0 (30):\penalty0 4456--4471,
  2012.

\bibitem[Crowther et~al.(2013)Crowther, Abrams, and Lambert]{crowther2013joint}
Crowther, M.~J., Abrams, K.~R., and Lambert, P.~C.
\newblock Joint modeling of longitudinal and survival data.
\newblock \emph{The Stata Journal}, 13\penalty0 (1):\penalty0 165--184, 2013.

\bibitem[Damianou \& Lawrence(2013)Damianou and Lawrence]{damianou2013deep}
Damianou, A. and Lawrence, N.~D.
\newblock Deep gaussian processes.
\newblock In \emph{Artificial Intelligence and Statistics}, pp.\  207--215,
  2013.

\bibitem[Dempsey et~al.(2017)Dempsey, Moreno, Scott, Dennis, Gustafson, Murphy,
  and Rehg]{dempsey2017isurvive}
Dempsey, W.~H., Moreno, A., Scott, C.~K., Dennis, M.~L., Gustafson, D.~H.,
  Murphy, S.~A., and Rehg, J.~M.
\newblock isurvive: An interpretable, event-time prediction model for mhealth.
\newblock In \emph{International Conference on Machine Learning}, pp.\
  970--979, 2017.

\bibitem[D{\"u}richen et~al.(2015)D{\"u}richen, Pimentel, Clifton, Schweikard,
  and Clifton]{durichen2015multitask}
D{\"u}richen, R., Pimentel, M.~A., Clifton, L., Schweikard, A., and Clifton,
  D.~A.
\newblock Multitask gaussian processes for multivariate physiological
  time-series analysis.
\newblock \emph{IEEE Transactions on Biomedical Engineering}, 62\penalty0
  (1):\penalty0 314--322, 2015.

\bibitem[Fern{\'a}ndez et~al.(2016)Fern{\'a}ndez, Rivera, and
  Teh]{fernandez2016gaussian}
Fern{\'a}ndez, T., Rivera, N., and Teh, Y.~W.
\newblock Gaussian processes for survival analysis.
\newblock In \emph{Advances in Neural Information Processing Systems}, pp.\
  5021--5029, 2016.

\bibitem[Gao et~al.(2015)Gao, Wang, Teti, Dornfeld, Kumara, Mori, and
  Helu]{gao2015cloud}
Gao, R., Wang, L., Teti, R., Dornfeld, D., Kumara, S., Mori, M., and Helu, M.
\newblock Cloud-enabled prognosis for manufacturing.
\newblock \emph{CIRP annals}, 64\penalty0 (2):\penalty0 749--772, 2015.

\bibitem[Gasmi et~al.(2003)Gasmi, Love, and Kahle]{gasmi2003general}
Gasmi, S., Love, C.~E., and Kahle, W.
\newblock A general repair, proportional-hazards, framework to model complex
  repairable systems.
\newblock \emph{IEEE Transactions on Reliability}, 52\penalty0 (1):\penalty0
  26--32, 2003.

\bibitem[Guarnizo \& {\'A}lvarez(2018)Guarnizo and
  {\'A}lvarez]{guarnizo2018fast}
Guarnizo, C. and {\'A}lvarez, M.~A.
\newblock Fast kernel approximations for latent force models and convolved
  multiple-output gaussian processes.
\newblock \emph{arXiv preprint arXiv:1805.07460}, 2018.

\bibitem[He et~al.(2015)He, Tu, Wang, Fu, and Yu]{he2015simultaneous}
He, Z., Tu, W., Wang, S., Fu, H., and Yu, Z.
\newblock Simultaneous variable selection for joint models of longitudinal and
  survival outcomes.
\newblock \emph{Biometrics}, 71\penalty0 (1):\penalty0 178--187, 2015.

\bibitem[Kalbfleisch \& Prentice(2011)Kalbfleisch and
  Prentice]{kalbfleisch2011statistical}
Kalbfleisch, J.~D. and Prentice, R.~L.
\newblock \emph{The statistical analysis of failure time data}, volume 360.
\newblock John Wiley \& Sons, 2011.

\bibitem[Kim \& Pavlovic(2018)Kim and Pavlovic]{Minyoung2018}
Kim, M. and Pavlovic, V.
\newblock Variational inference for gaussian process models for survival
  analysis.
\newblock \emph{Uncertainty in Artificial Intelligence}, 2018.

\bibitem[Kontar et~al.(2018{\natexlab{a}})Kontar, Zhou, Sankavaram, Du, and
  Zhang]{kontar2018nonparametrica}
Kontar, R., Zhou, S., Sankavaram, C., Du, X., and Zhang, Y.
\newblock Nonparametric-condition-based remaining useful life prediction
  incorporating external factors.
\newblock \emph{IEEE Transactions on Reliability}, 67\penalty0 (1):\penalty0
  41--52, 2018{\natexlab{a}}.

\bibitem[Kontar et~al.(2018{\natexlab{b}})Kontar, Zhou, Sankavaram, Du, and
  Zhang]{kontar2018nonparametricb}
Kontar, R., Zhou, S., Sankavaram, C., Du, X., and Zhang, Y.
\newblock Nonparametric modeling and prognosis of condition monitoring signals
  using multivariate gaussian convolution processes.
\newblock \emph{Technometrics}, 60\penalty0 (4):\penalty0 484--496,
  2018{\natexlab{b}}.

\bibitem[Liu et~al.(2013)Liu, Gebraeel, and Shi]{liu2013data}
Liu, K., Gebraeel, N.~Z., and Shi, J.
\newblock A data-level fusion model for developing composite health indices for
  degradation modeling and prognostic analysis.
\newblock \emph{IEEE Transactions on Automation Science and Engineering},
  10\penalty0 (3):\penalty0 652--664, 2013.

\bibitem[Mauff et~al.(2018)Mauff, Steyerberg, Kardys, Boersma, and
  Rizopoulos]{mauff2018joint}
Mauff, K., Steyerberg, E., Kardys, I., Boersma, E., and Rizopoulos, D.
\newblock Joint models with multiple longitudinal outcomes and a time-to-event
  outcome.
\newblock \emph{arXiv preprint arXiv:1808.07719}, 2018.

\bibitem[Moreno-Mu{\~n}oz et~al.(2018)Moreno-Mu{\~n}oz,
  Art{\'e}s-Rodr{\'\i}guez, and {\'A}lvarez]{moreno2018heterogeneous}
Moreno-Mu{\~n}oz, P., Art{\'e}s-Rodr{\'\i}guez, A., and {\'A}lvarez, M.~A.
\newblock Heterogeneous multi-output gaussian process prediction.
\newblock \emph{arXiv preprint arXiv:1805.07633}, 2018.

\bibitem[Pham et~al.(2012)Pham, Yang, and Nguyen]{pham2012machine}
Pham, H.~T., Yang, B., and Nguyen, T.~T.
\newblock Machine performance degradation assessment and remaining useful life
  prediction using proportional hazard model and support vector machine.
\newblock \emph{Mechanical Systems and Signal Processing}, 32:\penalty0
  320--330, 2012.

\bibitem[Proust-Lima et~al.(2014)Proust-Lima, S{\'e}ne, Taylor, and
  Jacqmin-Gadda]{proust2014joint}
Proust-Lima, C., S{\'e}ne, M., Taylor, J.~M., and Jacqmin-Gadda, H.
\newblock Joint latent class models for longitudinal and time-to-event data: A
  review.
\newblock \emph{Statistical methods in medical research}, 23\penalty0
  (1):\penalty0 74--90, 2014.

\bibitem[Rizopoulos(2011)]{rizopoulos2011dynamic}
Rizopoulos, D.
\newblock Dynamic predictions and prospective accuracy in joint models for
  longitudinal and time-to-event data.
\newblock \emph{Biometrics}, 67\penalty0 (3):\penalty0 819--829, 2011.

\bibitem[Rizopoulos(2012)]{rizopoulos2012joint}
Rizopoulos, D.
\newblock \emph{Joint models for longitudinal and time-to-event data: With
  applications in R}.
\newblock Chapman and Hall/CRC, 2012.

\bibitem[Rizopoulos et~al.(2017)Rizopoulos, Molenberghs, and
  Lesaffre]{rizopoulos2017dynamic}
Rizopoulos, D., Molenberghs, G., and Lesaffre, E.~M.
\newblock Dynamic predictions with time-dependent covariates in survival
  analysis using joint modeling and landmarking.
\newblock \emph{Biometrical Journal}, 59\penalty0 (6):\penalty0 1261--1276,
  2017.

\bibitem[Rosenberg(1995)]{rosenberg1995hazard}
Rosenberg, P.~S.
\newblock Hazard function estimation using b-splines.
\newblock \emph{Biometrics}, pp.\  874--887, 1995.

\bibitem[Ruppert(2002)]{ruppert2002selecting}
Ruppert, D.
\newblock Selecting the number of knots for penalized splines.
\newblock \emph{Journal of computational and graphical statistics}, 11\penalty0
  (4):\penalty0 735--757, 2002.

\bibitem[Saxena \& Goebel(2008)Saxena and Goebel]{saxena2008}
Saxena, A. and Goebel, K.
\newblock Turbofan engine degradation simulation data set, 2008.
\newblock data retrieved from NASA Ames Prognostics Data Repository,
  \url{https://ti.arc.nasa.gov/tech/dash/groups/pcoe/prognostic-data-repository/}.

\bibitem[Snelson \& Ghahramani(2006)Snelson and Ghahramani]{snelson2006sparse}
Snelson, E. and Ghahramani, Z.
\newblock Sparse gaussian processes using pseudo-inputs.
\newblock In \emph{Advances in neural information processing systems}, pp.\
  1257--1264, 2006.

\bibitem[Soleimani et~al.(2018)Soleimani, Hensman, and
  Saria]{soleimani2018scalable}
Soleimani, H., Hensman, J., and Saria, S.
\newblock Scalable joint models for reliable uncertainty-aware event
  prediction.
\newblock \emph{IEEE transactions on pattern analysis and machine
  intelligence}, 40\penalty0 (8):\penalty0 1948--1963, 2018.

\bibitem[Titsias \& Lawrence(2010)Titsias and Lawrence]{titsias2010bayesian}
Titsias, M. and Lawrence, N.~D.
\newblock Bayesian gaussian process latent variable model.
\newblock In \emph{Proceedings of the Thirteenth International Conference on
  Artificial Intelligence and Statistics}, pp.\  844--851, 2010.

\bibitem[Tsiatis et~al.(1995)Tsiatis, Degruttola, and
  Wulfsohn]{tsiatis1995modeling}
Tsiatis, A.~A., Degruttola, V., and Wulfsohn, M.~S.
\newblock Modeling the relationship of survival to longitudinal data measured
  with error. applications to survival and cd4 counts in patients with aids.
\newblock \emph{Journal of the American Statistical Association}, 90\penalty0
  (429):\penalty0 27--37, 1995.

\bibitem[Wu et~al.(2012)Wu, Liu, Yi, and Huang]{wu2012analysis}
Wu, L., Liu, W., Yi, G.~Y., and Huang, Y.
\newblock Analysis of longitudinal and survival data: joint modeling, inference
  methods, and issues.
\newblock \emph{Journal of Probability and Statistics}, 2012, 2012.

\bibitem[Wulfsohn \& Tsiatis(1997)Wulfsohn and Tsiatis]{wulfsohn1997joint}
Wulfsohn, M.~S. and Tsiatis, A.~A.
\newblock A joint model for survival and longitudinal data measured with error.
\newblock \emph{Biometrics}, pp.\  330--339, 1997.

\bibitem[Yan et~al.(2016)Yan, Liu, Zhang, and Shi]{yan2016multiple}
Yan, H., Liu, K., Zhang, X., and Shi, J.
\newblock Multiple sensor data fusion for degradation modeling and prognostics
  under multiple operational conditions.
\newblock \emph{IEEE Transactions on Reliability}, 65\penalty0 (3):\penalty0
  1416--1426, 2016.

\bibitem[Yu et~al.(2004)Yu, Law, Taylor, and Sandler]{yu2004joint}
Yu, M., Law, N.~J., Taylor, J.~M., and Sandler, H.~M.
\newblock Joint longitudinal-survival-cure models and their application to
  prostate cancer.
\newblock \emph{Statistica Sinica}, pp.\  835--862, 2004.

\bibitem[Zhao \& Sun(2016)Zhao and Sun]{zhao2016variational}
Zhao, J. and Sun, S.
\newblock Variational dependent multi-output gaussian process dynamical
  systems.
\newblock \emph{The Journal of Machine Learning Research}, 17\penalty0
  (1):\penalty0 4134--4169, 2016.

\bibitem[Zhou et~al.(2014)Zhou, Son, Zhou, Mao, and Salman]{zhou2014remaining}
Zhou, Q., Son, J., Zhou, S., Mao, X., and Salman, M.
\newblock Remaining useful life prediction of individual units subject to hard
  failure.
\newblock \emph{IIE Transactions}, 46\penalty0 (10):\penalty0 1017--1030, 2014.

\bibitem[Zhu et~al.(2012)Zhu, Ibrahim, Chi, and Tang]{zhu2012bayesian}
Zhu, H., Ibrahim, J.~G., Chi, Y., and Tang, N.
\newblock Bayesian influence measures for joint models for longitudinal and
  survival data.
\newblock \emph{Biometrics}, 68\penalty0 (3):\penalty0 954--964, 2012.

\end{thebibliography}
\bibliographystyle{icml2019}
\end{document}